\documentclass[final,3p]{elsarticle}

\usepackage{lineno}
\usepackage[colorlinks=true,breaklinks=true,pdftex]{hyperref}
\usepackage{amsmath}
\usepackage{amssymb,amsfonts}

\modulolinenumbers[1]

\journal{GMP 2025}

\biboptions{authoryear}

\begin{document}

\begin{frontmatter}

\title{Learning Contour-Guided 3D Face Reconstruction with Occlusions}

% USE FOR THE REVIEW PROCESS
% \author{Submission ID:1057}

%% ONLY USE FOR THE FINAL ACCEPTED PAPER SUBMISSION
\author[first]{Dapeng Zhao}
\ead{mirror1775@gmail.com}
% \author[last]{Yue Qi}
% \ead{LastAuthor@email.com}
% \cortext[cor]{Corresponding author}
\address[first]{Zhejiang Lab, Hangzhou,China}
% \address[last]{Institution and address of the last author}

\begin{abstract}
  3D face reconstruction is a versatile technology applied in various contexts. Deep learning methods, in particular, prove highly valuable in this domain due to their ability to generate high-quality 3D models. Nevertheless, most of these techniques require unobstructed and clear facial images as input. In response, we've developed a 3D face reconstruction system capable of performing effectively even when faced with obstructed views. Taking inspiration from generative face image inpainting, we've introduced a comprehensive method for crafting facial representations guided by their outlines. Our network comprises two components: one dedicated to image enhancement by removing obstructions and restoring missing facial features, while the other utilizes weak supervision to craft exceptionally detailed 3D models. Through extensive experimentation on standard 3D face reconstruction tasks, we've demonstrated the superiority of our method compared to existing ones that often fall short. Our results, based on experiments conducted with LFW databases, affirm the effectiveness of our approach.
\end{abstract}

\begin{keyword}
3D Reconstruction \sep Face Parsing \sep Face Recognition
\end{keyword}

\end{frontmatter}

% Comment out for final accepted paper submission
% \linenumbers

%%%%%%%%%%%%%%%%%%%%%%%%%%%%%%%%%%%%%%%%%%%%%%%%%%%%%%%%%%%%%%%%%%%%%
\section{Introduction}
Face reconstruction systems find extensive applications in specific scenarios, including face recognition~\cite{RN463} and digital entertainment~\cite{RN208}. Nonetheless, the majority of these systems require the input face photos to be unobstructed and depict the front of the face. In contrast to recent research efforts focused on enhancing facial resolution and fidelity~\cite{RN150}, 3D face reconstruction in occluded environments remains underexplored. Consequently, addressing the challenge of reconstructing faces in obstructed scenes is a pressing task.
\\Face images can be occluded by anything, such as palms, glasses, or leaves. Without any prior or weak supervision, it has been shown that we cannot simply infer the occluded face image. However, face image synthesis has achieved tremendous success in recent years due to the rapid development of Generative Adversarial Networks (GANs). State-of-the-art GAN techniques, such as Stylegan2~\cite{RN754}, can generate high-fidelity virtual face images that are sometimes even difficult to distinguish from real ones.
\\In this work, we investigated the identification of occluded areas and the removal of occlusions to synthesize a complete face from a human face. We used face parsing networks to identify the occluded areas and used a contour-based GAN to synthesize a complete 2D face. Due to the lack of 3D datasets, we utilized a weakly supervised deep learning framework to predict the coefficients of the 3D Morphable Model (3DMM). In the end, we obtained the final face model with complete facial features.
\\\textit{This research makes three contributions:}
\\$\bullet$\ We propose an algorithm that combines face analysis map and face contour map to generate face with full facial features.
\\$\bullet$\ We have improved the loss function of our 3D reconstruction framework for occluded scenes. Our results are more accurate than other state-of-the-art methods.
\\$\bullet$\ We propose a novel 3D face reconstruction method that can produce accurate models under occluded scenes.
\section{Related Work}
\subsection{Generic Face Reconstruction}
Deep learning-based methods~\cite{RN41} directly regress 3D Morphable Model (3DMM) coefficients from images. Some frameworks~\cite{RN581} have explored end-to-end convolutional neural network (CNN) architectures to regress 3DMM coefficients directly. However, most of these methods require that the input image is an unobstructed frontal face. In order to obtain paired 2D-3D data for supervised learning, these approaches typically require very deep CNNs, which can be difficult to train. Additionally, these approaches do not perform well when dealing with complex lighting, occlusion, and other in-the-wild conditions.
\subsection{Face Parsing}
Face parsing maps are commonly used as an essential intermediate representation for generative modeling~\cite{RN589}. They can be used as the conditional context for conditional face image synthesis. Image-to-image translation models can learn the mapping from parsing maps to realistic RGB images~\cite{RN257}. A practical face parsing solution should directly predict per-pixel semantic labels across the entire face image.
To obtain accurate parsing maps, Wei \textit{et al.}~\cite{RN615} developed a novel method for regulating receptive fields with superior regulation ability in parsing networks. MaskGan~\cite{RN311} provided a labeled face parsing dataset. Zhao \textit{et al.}~\cite{RN616} developed a model that explored how to combine the fully convolutional network and super-pixels data.
To address the problem of limited access to global image information, some methods~\cite{RN621} have introduced the transformer component. Semantic layouts provide rough guidance of the location and appearance of objects, which further facilitates training.
\subsection{Face Image Synthesis}
Due to the rapid development of convolutional neural networks, deep pixel-level face generation has shown extraordinary capabilities. Recently, several methods have been introduced that provide additional information before face image synthesis.
One of the first deep learning methods designed for image inpainting is context encoder~\cite{RN290}, which uses an encoder-decoder framework. However, the network performs poorly on human faces. Following this work, Dolhansky \textit{et al.}~\cite{RN618} demonstrated the importance of exemplar data for inpainting. However, this method only focuses on filling in missing eye regions of frontal faces, so it does not generalize well.
In subsequent work, a pre-trained VGG network was used to improve the output of the context encoder by minimizing the feature differences between the input image and the inpainted image~\cite{RN806}.
\\
In the face synthesis task, the encoder converts an image with the concerned regions into a low-dimensional feature space. The decoder then builds the output image. However, due to the data bottleneck in the channel-wise fully connected layer, the restored regions of the output image often contain visual artifacts and appear blurry.
Iizuka \textit{et al.}~\cite{RN805} solved this problem by replacing the fully connected layer with a series of dilated convolutional layers. Yu \textit{et al.}~\cite{RN751} proposed a two-step image synthesis method. In the first step, it estimates the missing area in a coarse manner. The refinement network then uses an attention mechanism to sharpen the result by looking for a collection of background patches with the highest similarity to the coarse estimate.
\section{Proposed Approach}
\begin{figure*}[htb]
    % \vspace{-1.0em}
    \centering
\includegraphics[width=0.70\textwidth]{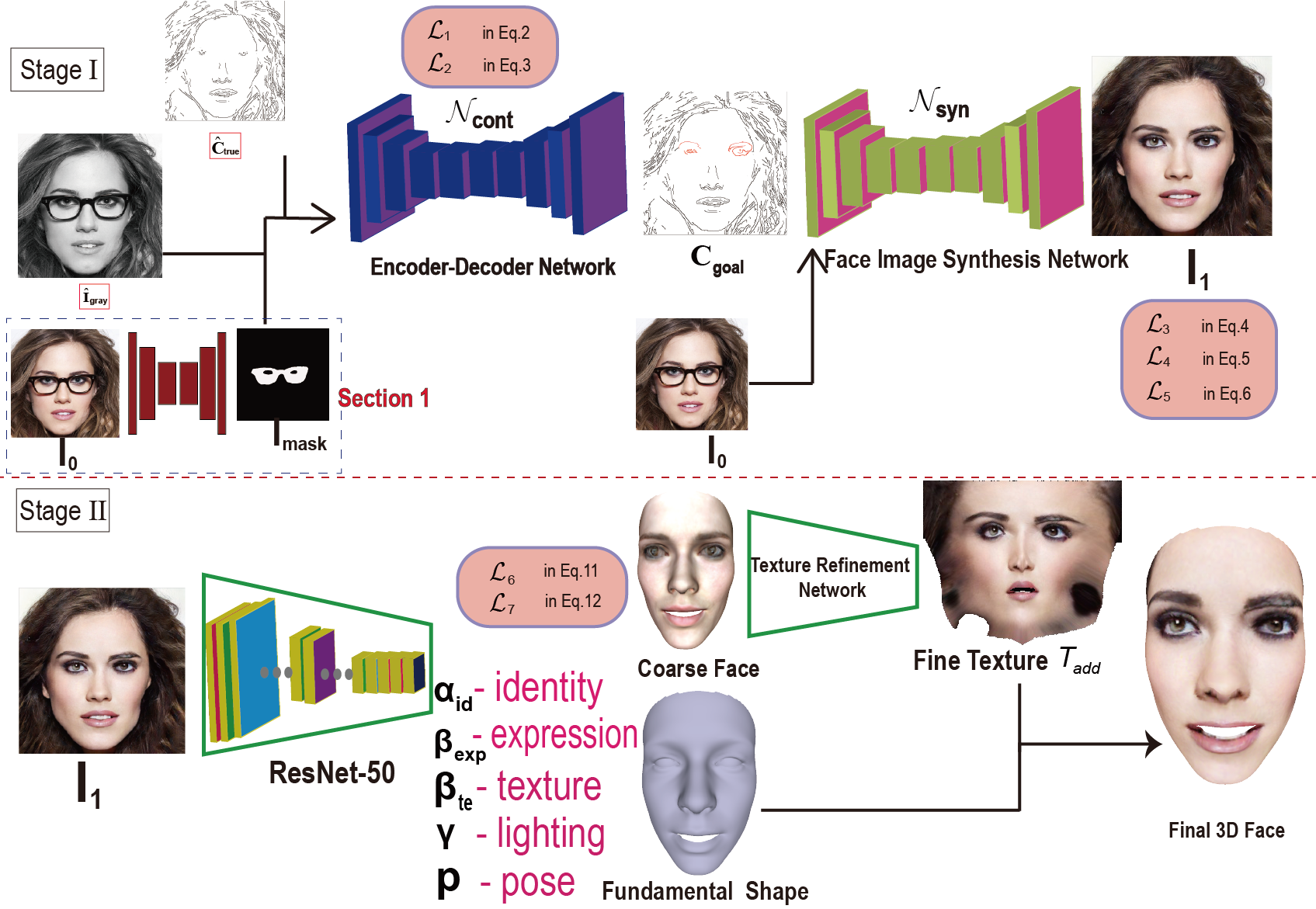}
\caption{Method overview. See related sections for details.} \label{fig:overall}
% \vspace{-4.0em}
\end{figure*}
We propose a 3D face reconstruction method based on a single image (as shown in Figure~\ref{fig:overall}). The method consists of two stages.  The first stage, face image synthesis under occluded scenes, synthesizes a complete face image from an occluded image. The second stage, a 3D face reconstruction network based on unobstructed frontal images, reconstructs a 3D face model from the synthesized complete face image. Our goal is to realize 3D face reconstruction under occluded scenes using this framework.Given a source image ${{\mathbf{I}}_{\mathbf{input}}}\in {{\mathbb{R}}^{H\times W\times 3}}$ with obstructions on the face, we obtain the final face model.
\subsection{How to Obtain the Face Mask}
\begin{figure}[h]
    % \vspace{-1.0em}
    \centering
\includegraphics[width=0.35\textwidth]{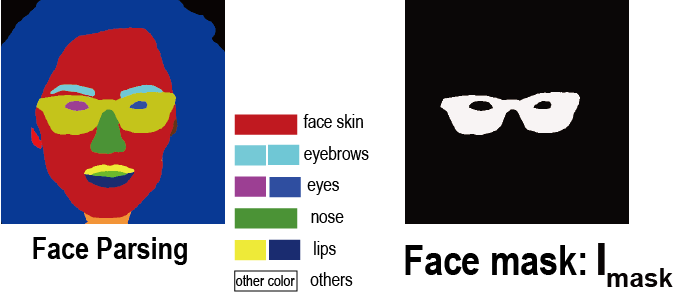}
\caption{The face mask we define, indicating the position of the occlusions.} \label{fig:mask}
% \vspace{-4.0em}
\end{figure}
As show in Section 1 of Figure~\ref{fig:overall} and Figure~\ref{fig:mask}, 
we define face mask image ${{\mathbf{I}}_{\mathbf{mask}}}\in {{\mathbb{R}}^{H\times W\times 1}}$ ($1$ for the occluded occlusions,$0$ for background) indicating the position of the occlusions. Our face 
mask ${{\mathbf{I}}_{\mathbf{mask}}}$ and traditional face parsing map are closely related. We need to separate the facial features, and we can reveal the area of the occlusions. Please notice that, in our work, we assumed that facial features only include only ﬁve parts, including facial skin, eyebrows, eyes, nose and lips. Inspired by the method of CelebAMask-HQ~\cite{RN311} , given an input face image ${{\mathbf{I}}_{\mathbf{0}}}\in {{\mathbb{R}}^{H\times W\times 3}}$  under extreme scenes, we leveraged an 
encoder-decoder module ${{\Omega }_{m}}$  based on U-Net to obtain the face mask.
\subsection{Complete Face Image Generation}
Our image synthesis module is guided by the contour (some approaches called \textit{edge}). First, we need to predict the contours ${{\mathbf{C}}_{\mathbf{syn}}}\in {{\mathbb{R}}^{H\times W\times 1}}$ of facial features in the occluded area. We assume that the final unobstructed face image is ${{\mathbf{I}}_{\mathbf{1}}}\in {{\mathbb{R}}^{H\times W\times 3}}$ and the ground truth image without obstruction ${{\mathbf{I}}_{\mathbf{true}}}\in {{\mathbb{R}}^{H\times W\times 3}}$. In the training set, the corresponding complete contour image and gray image are ${{\mathbf{C}}_{\mathbf{true}}}\in {{\mathbb{R}}^{H\times W\times 1}}$ and ${{\mathbf{I}}_{\mathbf{gray}}}\in {{\mathbb{R}}^{H\times W\times 1}}$. We trained the contour generator ${{\mathcal{N}}_{cont}}$ to predict the contour map for the occluded region
\begin{equation}
\left\{ \begin{aligned}
    & {{\mathbf{C}}_{\mathbf{syn}}}={{\mathcal{N}}_{cont}}\left( {{{\mathbf{\hat{I}}}}_{\mathbf{gray}}},{{\mathbf{I}}_{\mathbf{mask}}},{{{\mathbf{\hat{C}}}}_{\mathbf{true}}} \right) \\ 
   & {{{\mathbf{\hat{I}}}}_{\mathbf{gray}}}\text{=}{{\mathbf{I}}_{\mathbf{gray}}}\odot \left( 1-{{\mathbf{I}}_{\mathbf{mask}}} \right) \\ 
   & {{{\mathbf{\hat{C}}}}_{\mathbf{ture}}}\text{=}{{\mathbf{C}}_{\mathbf{ture}}}\odot \left( 1-{{\mathbf{I}}_{\mathbf{mask}}} \right) \\ 
  \end{aligned} \right.
\end{equation}
\ \ where ${{\mathbf{\hat{I}}}_{\mathbf{gray}}}$ denotes masked grayscale image,${{\mathbf{\hat{C}}}_{\mathbf{ture}}}$ denotes masked contour image and $\odot $ denotes the Hadamard product.
\\We trained the discriminator of the module ${{\mathcal{N}}_{cont}}$ to predict which 
of ${{\mathbf{C}}_{\mathbf{syn}}}$ and ${{\mathbf{C}}_{\mathbf{ture}}}$ is true contour map and which is false contour map. The adversarial loss is defined as
\begin{equation}
    \begin{aligned}
        & {{\mathcal{L}}_{1}}={{\mathbb{E}}_{\left( {{\mathbf{C}}_{\mathbf{ture}}},{{\mathbf{I}}_{\mathbf{gray}}} \right)}}\left[ \log {{D}_{1}}\left( {{\mathbf{C}}_{\mathbf{true}}},{{\mathbf{I}}_{\mathbf{gray}}} \right) \right] \\ 
       & +{{\mathbb{E}}_{\left( {{\mathbf{I}}_{\mathbf{gray}}} \right)}}\log \left[ 1-{{D}_{1}}\left( {{\mathbf{C}}_{\mathbf{syn}}},{{\mathbf{I}}_{\mathbf{gray}}} \right) \right] \\ 
      \end{aligned}
\end{equation}
\ \ In addition, we compare the feature activation maps of the discriminator. We set the face feature matching loss as
\begin{equation}
    {{\mathcal{L}}_{2}}=\mathbb{E}\left[ \sum\limits_{\text{i}=1}^{\text{K}}{\frac{1}{{{N}_{i}}}}\left\| D_{1}^{(i)}({{\mathbf{C}}_{\mathbf{true}}})-D_{1}^{(i)}({{\mathbf{C}}_{\mathbf{syn}}}) \right\| \right]
\end{equation}
\ \ where ${{N}_{i}}$ is the number of elements in the $i$th activation layer,$K$ is the ﬁnal convolution layer of the discriminator and $D_{1}^{(i)}$ is the activation in the $i$th layer of the discriminator.
\\After obtaining the complete contour map, we design ${{\mathcal{N}}_{syn}}$ to generate the complete face image ${{\mathbf{I}}_{\mathbf{1}}}$. The complete contour map ${{\mathbf{C}}_{\mathbf{goal}}}$ is formed by adding ${{\mathbf{C}}_{\mathbf{syn}}}$  and ${{\mathbf{C}}_{\mathbf{true}}}$, which follows ${{\mathbf{C}}_{\mathbf{goal}}}\text{=}{{\mathbf{C}}_{\mathbf{true}}}\odot (1-{{\mathbf{I}}_{\mathbf{mask}}})+{{\mathbf{C}}_{\mathbf{syn}}}\odot {{\mathbf{I}}_{\mathbf{mask}}}$ . In the map ${{\mathbf{C}}_{\mathbf{goal}}}$, we can see the contours of all facial features, especially the occluded areas. In addition, 
we set ${{\mathbf{\hat{I}}}_{\mathbf{true}}}\in {{\mathbb{R}}^{H\times W\times 3}}$ to be an incomplete face picture, which follows ${{\mathbf{\hat{I}}}_{\mathbf{true}}}\text{=}{{\mathbf{I}}_{\mathbf{true}}}\odot (1-{{\mathbf{I}}_{\mathbf{mask}}})$.So, we utilize ${{\mathcal{N}}_{syn}}$  to get the final complete face image ${{\mathbf{I}}_{\mathbf{1}}}$ , with occluded regions recovered, which follows ${{\mathbf{I}}_{\mathbf{1}}}\text{=}{{\mathcal{N}}_{syn}}({{\mathbf{\hat{I}}}_{\mathbf{true}}},{{\mathbf{C}}_{\mathbf{goal}}})$.
\\We trained the module ${{\mathcal{N}}_{syn}}$  to predict the final complete face image ${{\mathbf{I}}_{\mathbf{1}}}$ over a joint loss. The adversarial loss is defined as
\begin{equation}
    \begin{aligned}
        & {{\mathcal{L}}_{3}}={{\mathbb{E}}_{\left( {{\mathbf{I}}_{\mathbf{true}}},{{\mathbf{C}}_{\mathbf{fina}}} \right)}}\left[ \log {{D}_{2}}\left( {{\mathbf{I}}_{\mathbf{true}}},{{\mathbf{C}}_{\mathbf{goal}}} \right) \right] \\ 
       & +{{\mathbb{E}}_{\left( {{\mathbf{C}}_{\mathbf{fina}}} \right)}}\log \left[ 1-{{D}_{2}}\left( {{\mathbf{I}}_{\mathbf{1}}},{{\mathbf{C}}_{\mathbf{goal}}} \right) \right] \\ 
      \end{aligned}
\end{equation}
\ \ The per-pixel loss is deﬁned as follows
\begin{equation}
    {{\mathcal{L}}_{4}}=\frac{1}{{{\text{S}}_{m}}}{{\left\| {{\mathbf{I}}_{\mathbf{1}}}-{{\mathbf{I}}_{\mathbf{true}}} \right\|}_{1}}
\end{equation}
\ \ which ${{S}_{m}}$ denotes the size of the face mask ${{\mathbf{I}}_{\mathbf{mask}}}$ and ${{\left\| \cdot  \right\|}_{1}}$ denotes the ${{L}_{1}}$ norm. Notice that we use the mask size ${{S}_{m}}$ as the denominator to adjust the penalty.
\\The style loss~\cite{51} computes the style distance between two face images as follows 
\begin{equation}
    \begin{aligned}
        & {{\mathcal{L}}_{5}}= \\ 
       & {{\sum\limits_{\text{n}}{\frac{1}{{{Q}_{n}}\times {{Q}_{n}}}\left\| \frac{{{G}_{n}}({{\mathbf{I}}_{\mathbf{1}}}\odot (1-{{\mathbf{I}}_{\mathbf{mask}}}))-{{G}_{n}}({{{\mathbf{\hat{I}}}}_{\mathbf{true}}})}{{{Q}_{n}}\times {{H}_{n}}\times {{W}_{n}}} \right\|}}_{1}} \\ 
      \end{aligned}
\end{equation}
\ \ where ${{G}_{\text{n}}}\text{(x)=}{{\varphi }_{n}}{{(x)}^{T}}{{\varphi }_{n}}(x)$ denotes the Gram Matrix corresponding to ${{\varphi }_{n}}(x)$,${{\varphi }_{n}}(\cdot )$ denotes 
the ${{Q}_{n}}$ feature maps with the size ${{H}_{n}}\times {{W}_{n}}$ of the $n$-th layer.
\ \ In summary, the contour generator network ${{\mathcal{N}}_{cont}}$  was trained with an objective comprised of an adversarial loss and feature-matching loss
\begin{equation}
    \underset{{{G}_{1}}}{\mathop{\text{min}}}\,\underset{{{D}_{1}}}{\mathop{\text{max}}}\,{{\mathcal{L}}_{{{G}_{1}}}}=\underset{{{G}_{1}}}{\mathop{\text{min}}}\,\left( {{\lambda }_{1}}\underset{{{D}_{1}}}{\mathop{\text{max}}}\,{{\mathcal{L}}_{1}}+{{\lambda }_{2}}{{\mathcal{L}}_{2}} \right)
\end{equation}
\ \ The total loss function of ${{\mathcal{N}}_{syn}}$ follows
\begin{equation}
    \underset{{{G}_{2}}}{\mathop{\text{min}}}\,\underset{{{D}_{2}}}{\mathop{\text{max}}}\,{{\mathcal{L}}_{{{G}_{2}}}}={{\lambda }_{3}}\underset{{{D}_{2}}}{\mathop{\text{max}}}\,{{\mathcal{L}}_{3}}+{{\lambda }_{4}}{{\mathcal{L}}_{4}}+{{\lambda }_{5}}{{\mathcal{L}}_{5}}
\end{equation}
here, we set ${{\lambda }_{1}}\text{=}1$ ,${{\lambda }_{2}}\text{=}11.5$ ,${{\lambda }_{3}}\text{=0}.1$ , ${{\lambda }_{4}}\text{=}1$ and ${{\lambda }_{5}}\text{=250}$  respectively.
\subsection{3D Model Reconstruction}
Since the face is a very standardized figure, we use a human face template to construct the basic shape (here, we adopt 3DMM)~\cite{RN45}. Commonly, we describe single people's face with 
PCA~\cite{RN463}, where shape and texture are separated:
\begin{equation}
    \left\{ \begin{aligned}
        & \mathbf{S}=\overline{\mathbf{S}}+{{\mathbf{A}}_{\mathbf{id}}}{{\mathbf{\alpha }}_{\mathbf{id}}}+{{\mathbf{B}}_{\mathbf{exp}}}{{\mathbf{\beta }}_{\mathbf{exp}}} \\ 
       & \mathbf{T}=\overline{\mathbf{T}}+{{\mathbf{B}}_{\mathbf{te}}}{{\mathbf{\beta }}_{\mathbf{te}}} \\ 
      \end{aligned} \right.
\end{equation}
where $\overline{\mathbf{S}}$ and $\overline{\mathbf{T}}$ denote the average shape and texture, ${{\mathbf{A}}_{\mathbf{id}}}$ , ${{\mathbf{B}}_{\mathbf{exp}}}$ and ${{\mathbf{B}}_{\mathbf{te}}}$ denote the PCA basis of shape (identity and expression) and texture.
${{\mathbf{\alpha }}_{\mathbf{id}}}\in {{\mathbb{R}}^{80}}$ , ${{\mathbf{\beta }}_{\mathbf{exp}}}\in {{\mathbb{R}}^{64}}$ and ${{\mathbf{\beta }}_{\mathbf{te}}}\in {{\mathbb{R}}^{80}}$ are the corresponding 3DMM coefficient vectors. After the 3D face is reconstructed, it can be projected onto the image plane with the perspective projection:
\begin{equation}
    {{\mathbf{P}}_{\mathbf{2d}}}=k*{{\mathbf{P}}_{\mathbf{p}}}*\mathbf{R}*{{\mathbf{S}}_{\mathbf{mod}}}+{{\mathbf{t}}_{\mathbf{2d}}}
\end{equation}
where ${{\mathbf{P}}_{\mathbf{2d}}}$ denotes the projection function that turned the 3D model into 2D face positions, $k$ denotes the scale factor, ${{\mathbf{P}}_{\mathbf{p}}}$ denotes the projection matrix, $\mathbf{R}\in SO(3)$ denotes the rotation matrix and ${{\mathbf{t}}_{\mathbf{2d}}}\in {{\mathbb{R}}^{3}}$  denotes the translation vector.
\\We approximated the scene illumination with Spherical Harmonics (SH)~\cite{RN642} 
parameterized by coefﬁcient vector $\gamma \in {{\mathbb{R}}^{9}}$ . Overall, the unknown vector parameters can be formulated ${{\mathbf{V}}_{\mathbf{x}}}=({{\mathbf{\alpha }}_{\mathbf{id}}},{{\mathbf{\beta }}_{\mathbf{exp}}},{{\mathbf{\beta }}_{\mathbf{te}}},\mathbf{\gamma },\mathbf{p})\in {{\mathbb{R}}^{239}}$ , 
where $\mathbf{p}\in {{\mathbb{R}}^{6}}=\{\mathbf{pitch},\mathbf{yaw},\mathbf{roll},k,{{\mathbf{t}}_{\mathbf{2D}}}\}$  denotes face poses. Here, we used a revised 
ResNet-50~\cite{RN662} network to predict ${{\mathbf{V}}_{\mathbf{x}}}$ . 
\quad The corresponding loss function consists of two parts:pixel-wise loss and face feature loss.
\\\qquad \textit{Per-pixel Loss}. The pixel loss function minimizes the difference between the input image $\mathbf{I}_{\mathbf{fina}}^{(j)}$ and the output image $\mathbf{I}_{\mathbf{y}}^{(j)}$ . The rendering layer renders back a 
rendered image $\mathbf{I}_{\mathbf{y}}^{(j)}$ to compare with the image $\mathbf{I}_{\mathbf{1}}^{(j)}$ . We compute the per-pixel loss with:
\begin{equation}
    {{\mathcal{L}}_{6}}={{\left\| \mathbf{I}_{\mathbf{1}}^{(j)}-\mathbf{I}_{\mathbf{y}}^{(j)} \right\|}_{2}}
\end{equation}
where $j$ denotes pixel index and ${{\left\| \cdot  \right\|}_{2}}$ denotes the ${{L}_{2}}$ norm.
\\\textit{Face Features Loss}. We introduce a loss function at the face recognition level to reduce the difference between the 3D model of the face and the 2D image. The loss function computes the feature diﬀerence between the input image $\mathbf{I}_{\mathbf{1}}^{{}}$ and rendered image $\mathbf{I}_{\mathbf{y}}^{{}}$ . We deﬁne the loss as a cosine distance:
\begin{equation}
    {{\mathcal{L}}_{7}}=1-\frac{<G(\mathbf{I}_{\mathbf{1}}^{{}}),G(\mathbf{I}_{\mathbf{y}}^{{}})>}{\left\| G(\mathbf{I}_{\mathbf{1}}^{{}}) \right\|\cdot \left\| G(\mathbf{I}_{\mathbf{y}}^{{}} \right\|}
\end{equation}
where ${G}(\cdot )$  denotes the feature extraction function by FaceNet~\cite{RN709},
$<\cdot ,\cdot >$ denotes the inner product.
\\In summary, we used the loss function ${{\mathcal{L}}_{3D}}$  to reconstruct the basic shape of the face.
 We set ${{\mathcal{L}}_{3D}}={{\lambda }_{6}}{{\mathcal{L}}_{6}}+{{\lambda }_{7}}{{\mathcal{L}}_{7}}$ , where ${{\lambda }_{6}}\text{=}1.4$  and  ${{\lambda }_{7}}\text{=0}\text{.25}$ respectively in all our experiments. We then used a coarse-to-fine graph convolutional network based on the frameworks of Lin \textit{et al.}~\cite{RN150} for producing the 
 fine texture ${{T}_{add}}$ .
\section{Experimental Results}
% \subsection{Qualitative Comparisons with Recent Works}
\begin{figure}[htb]
    % \vspace{-1.0em}
    \centering
\includegraphics[width=0.45\textwidth]{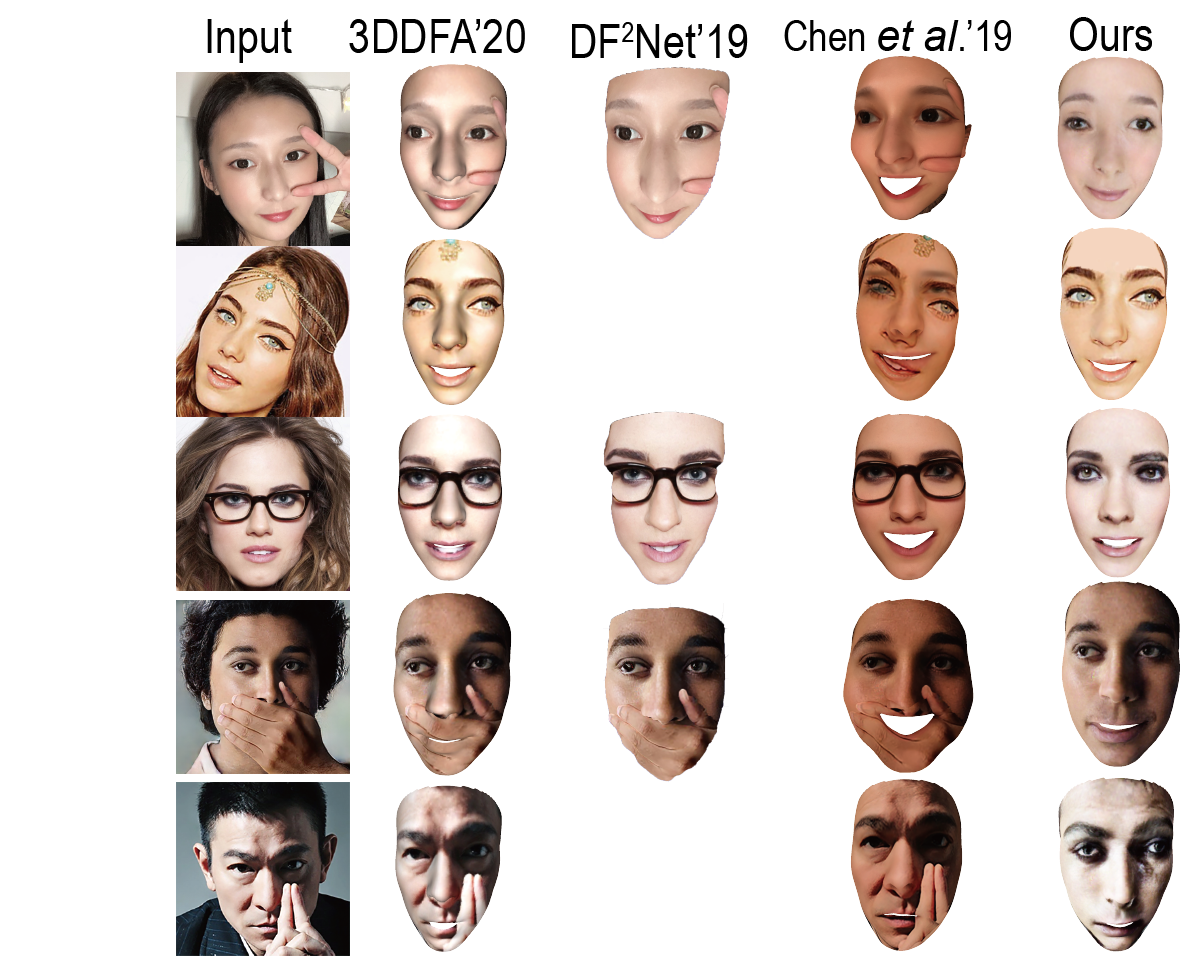}
\caption{ Comparison of qualitative results. Baseline methods from left to right: 3DDFA, DF2Net, Chen \textit{et al.} , and our method. The blank area means that the corresponding method does not work.} \label{bijiaotu}
% \vspace{-4.0em}
\end{figure}
Figire~\ref{bijiaotu} shows our experimental results compared with the others~\cite{RN251}. The result shows that our method is far superior to other frameworks. Our 3D reconstruction method can handle occluded scenes, such as palms, chains on the forehead, and glasses. Other frameworks can not handle occlusions well; they are more focused on the generation of high-definition textures.
\section{Conclusions}
In this study, we present an efficient 3D face reconstruction framework capable of functioning effectively in occluded environments. Given the wealth of domain knowledge and extensive prior research in the realm of human faces, we readily acquired relevant datasets. Our approach excels in estimating facial details even in areas where the face is obscured, such as by hands, forehead accessories, or glasses. Extensive experiments have conclusively demonstrated that our method significantly outperforms previous approaches in both accuracy and robustness.
%%%%%%%%%%%%%%%%%%%%%%%%%%%%%%%%%%%%%%%%%%%%%%%%%%%%%%%%%%%%%%%%%%%%%
%\section*{References} % needed on some systems
\bibliography{./mybibfile.bib}
\end{document}